\RequirePackage{amsmath}
\documentclass[runningheads]{llncs}
  \usepackage{graphicx}
  \usepackage{lmodern}
  \usepackage[utf8]{inputenc}
  \usepackage[T1]{fontenc}
  \usepackage{graphicx}
  \usepackage{algorithmicx}
  \usepackage{algorithm}
  \usepackage{algpseudocode} 
  \usepackage{todonotes}%
  \usepackage{bm}
  \usepackage{amsmath,amssymb,dsfont}
  \usepackage{mathtools}
  \usepackage{hyperref}

\title{A Linear Constrained Optimization Benchmark For Probabilistic Search Algorithms:}
\subtitle{The Rotated Klee-Minty Problem}
\titlerunning{A Linear Constrained Optimization Benchmark}

\author{Michael~Hellwig\orcidID{0000-0002-6731-8166} \and Hans-Georg~Beyer\orcidID{0000-0002-7455-8686}}

\authorrunning{M. Hellwig and H.-G. Beyer}

\institute{Vorarlberg University of Applied Sciences, Research Centre PPE,\\ Campus V, Hochschulstraße 1, 6850 Dornbirn, Austria.
\email{\{michael.hellwig,hans-georg.beyer\}@fhv.at}\\
\url{https://homepages.fhv.at/hemi/} \\ \url{https://homepages.fhv.at/hgb/}
}

\begin{document}

\maketitle              
\begin{abstract}
The development, assessment, and comparison of randomized search algorithms heavily rely on benchmarking. Regarding the domain of constrained optimization, the number of currently available benchmark environments bears no relation to the number of distinct problem features. The present paper advances a proposal of a scalable linear constrained optimization problem that is suitable for benchmarking Evolutionary Algorithms. By comparing two recent EA variants, the linear benchmarking environment is demonstrated.
\end{abstract}
\section{Introduction}

Benchmark environments establish well-defined experimental settings that aim at providing reproducible and comparable algorithmic results.
They are essential for the assessment and the comparison of contemporary algorithms. 
Benchmarking also is important for the development of new algorithmic ideas.
This is particularly true in the field of Evolutionary Algorithms (EA) for real-valued constrained optimization, where the theoretical background is comparably scarce.

Regarding EA benchmarks, the CEC competitions on constrained real-para\-meter optimization~\cite{CEC2006,CEC2010,CEC2017} introduced specific constrained test environments (usually referred to as constrained CEC benchmarks). 
The corresponding benchmark definitions supply a collection of mainly nonlinear objective functions that are constrained by various numbers of equality, inequality, and box-constraints. 
When considering real-valued constrained optimization problems, the CEC function sets represent the most frequently used benchmarking environment for contemporary EA.
Recently, a COCO branch for constrained black-box optimization benchmark problems~\cite{CocoCode} (BBOB-constrained\footnote{The code related to the BBOB-constrained suite under development is available in the \texttt{development} branch on \href{http://github.com/numbbo/coco/development}{http://github.com/numbbo/coco/development}.}) is developing. The BBOB-constrained test suite is a progression of the unconstrained COCO framework towards constrained benchmarks. However, it currently takes into account only a limited number of objective functions as well as almost linear inequality constraints of scalable quantity. 

Compared to the number of test problems currently used in constrained benchmark environments, the domain of real-valued constrained optimization problems is considerably larger. 
Constrained real-valued optimization problems may differ with respect to a multitude of features (and their combinations), including but not restricted to the number and type of constraints, the analytical structure of objective function, and the characteristics of the feasible region in the search space. Although first investigations exist~\cite{mezura2004makes}, it is not conclusively determined which features are making a constrained optimization problem hard.
Among the collections of constrained test problems available~\cite{NeumaierGlobalOpt}, the 
CEC and COCO benchmarks basically represent the two most elaborated benchmarking environments for EA~\cite{HellwigB2018b}. 
Considering that the EA development for constraint optimization tasks will further rely on suitable benchmarks and remembering the \emph{no free lunch} theorem~\cite{NFL1997}, the need for benchmark definitions that take into account consistent subgroups of conceivable problems is evident.

This paper presents a supplementary benchmark proposal. By providing hard but strictly linear constrained optimization problems that are scalable with respect to the problem dimension, it differs from the CEC and COCO environments.
The benchmark is constructed on the basis of the Klee-Minty polytope. It represents a unit hypercube of variable dimension with perturbed vertices~\cite{klee1970good}. The inside of the cube represents the feasible region of the constrained problem. The corresponding objective function is constructed in such a way that the classical Simplex algorithm yields an exponential worst-case running time, i.e. it can be considered hard with respect to computational complexity.

Considering the number of sophisticated deterministic approaches available, taking into account linear optimization problems for EA benchmarking might appear questionable. 
However, many purpose-built linear optimization algorithms show poor performance on the Klee-Minty problem. For example, other basis-exchange pivoting algorithms and even interior-point algorithms exhibit severe problems in this environment~\cite{IPM2,IPM1}. Contrary, some EA variants suited for generally constrained problems are able to obtain a similarly good or even better precision (cf. Tables~\ref{KMtab1} and~\ref{KMtab2}). In this regard, the Klee-Minty problem serves for demonstrating the applicability of EA to linear constrained black-box optimization problems. In~\cite{Spettel2018}, a Klee-Minty problem representation has already been used to assess the suitability of a custom-built Covariance Matrix Self-Adaptation Evolution Strategy for linearly constrained problems. The study substantiated a certain need for benchmarking functions suitable for testing EA that solely deal with linear constraints.

The present paper advances the Klee-Minty problem with respect to the following aspects.
We introduce an upstream motion in the search space that 
relocates the optimal solution which usually is placed on the axes.
This kind of location may present a bias towards Coordinate Search algorithms or box-constraint handling approaches.
The modified Klee-Minty problem is motivated in detail in Section~\ref{kleeminty}.
Moreover, in Section~\ref{convention} basic benchmarking conventions are proposed in order to provide a thorough basis for reproducible and comparable benchmarking tests.
The suggestion of a comprehensive presentation style and a comparison methodology for algorithm assessment are specified in Section~\ref{presentation}.
For demonstration purposes, two recent EA variants for constrained black-box optimization that proved successful in the context of the CEC competitions~\cite{CEC2017} are tested. 
The paper concludes with the discussion of currently unresolved issues and the suggestion of future development directions in Section~\ref{concl}.
	
\section{The rotated Klee-Minty problem}
	\label{kleeminty}
	
	The Klee-Minty cube (named after Victor Klee and George J. Minty) is a unit hypercube 
	of variable dimension with perturbed corners~\cite{klee1970good}.
	The inside of the cube represents the feasible region of a linear optimization problem which is referred to as the Klee-Minty problem. The corresponding objective function is constructed in such a way that the Simplex algorithm visits all the corners in the worst case and thus its worst-case runtime is exponential.

	Linear optimization test problems are inadequately represented in the context of EA benchmarking, as EAs usually cannot compete with custom-build linear solvers. However, the Klee-minty problem represents a reasonable hard linear problem that is suitable to present the potential of EAs.
	In order to remove undesired problem characteristics with respect to the location of the optimal solution and the orientation of the feasible region, 
	the Klee-Minty is modified by application of a transformation. 
	
	The introduction of a set of rotated Klee-Minty problems represents a benchmark proposal for reasonably hard linear optimization problems.
	The section provides a first suggestion in the style of the CEC benchmarks which is supported by ECDF performance plots~\cite{MoreWild2009}. 
	A modified Klee-Minty cube representation~\cite{deza2006central}\footnote{We omit use of the redundant constraints introduced in~\cite{deza2006central}. That is, $\bm{h}$ is considered to be an array of all-zeros for our Klee-Minty representation.} is considered to build the basis of the proposed linear benchmarks
	\begin{equation}
	  \begin{split}
	    \min_{\bm{y} \in \mathbb{R}^N} \: &\: \bm{c}^\top\bm{y}  \\
	    \textrm{s.t.} \: & \:  \bm{A}\bm{y} \leq \bm{b}, \\
			   &  \:  \check{\bm{y}} \leq \bm{y} \leq \hat{\bm{y}}. 
	  \end{split}
	\label{KMopt}
	\end{equation}
	The matrix $\bm{A}$ and the right-hand side vector $\bm{b}$ are defined as
	\begin{equation}
	  \bm{A} = \begin{pmatrix}
	             \bm{A}_1 \\
	             \bm{A}_2
	           \end{pmatrix} \in \mathds{R}^{2N \times N}, \qquad 
	        \bm{b} = \begin{pmatrix}
	                 \bm{1} \\
	                 \bm{0}
	               \end{pmatrix} \in \mathds{R}^{2N \times 1}.	               
	\end{equation}
	where $\bm{1}$ and $\bm{0}$ represent vectors of all ones, and all zeros, respectively. The $N\times N$ matrices $\bm{A}_1$ and $\bm{A}_2$ are defined as follows
	\begin{equation}
	  \bm{A}_1 = \begin{pmatrix}
	             1 & 0 & \dots & 0 & 0\\
	             \epsilon & 1  & \dots & 0 & 0 \\
	             \vdots & \epsilon & \ddots & \vdots & \vdots \\
	             \vdots & \vdots  & \ddots& \ddots & 0\\[1ex]
	             0 & 0&  \dots&   \epsilon & 1 
	           \end{pmatrix} \quad \textrm{and} \quad \bm{A}_2 =  \begin{pmatrix}
	             -1 & 0 & \dots & 0 & 0\\
	             \epsilon & -1  & \dots & 0 & 0 \\
	             \vdots & \epsilon & \ddots & \vdots & \vdots \\
	             \vdots & \vdots  & \ddots& \ddots & 0\\[1ex]
	             0 & 0&  \dots&   \epsilon & -1 
	           \end{pmatrix} .
	\end{equation}
	The parameter $0<\epsilon\leq1/3$ governs the perturbation of the unit cube. It is set to $\epsilon = 1/10$ to obtain problems of reasonable complexity. 
	Notice that, both matrices only differ with respect to the sign of their diagonal elements.
	
	While the above problem formulation strictly bounds the feasible region, we specify lower bounds $\check{\bm{y}}$ and upper bounds $\hat{\bm{y}}$ for the parameter vector components
	\begin{equation}
	 \check{\bm{y}} = \bm{0}\in \mathds{R}^N \quad \textrm{and} \quad  \hat{\bm{y}} = 5N^3\cdot\bm{1} \in \mathds{R}^N.
	\end{equation}
	When considering LP solvers that search exclusively inside the feasible region or on its borders (like the Simplex algorithm or interior point methods) the introduction of box constraints appears redundant. However, having in mind EA variants for constrained black-box optimization that move through infeasible regions of the search space, the box-constraints can be used to represent the domain of eligible input values\footnote{The Klee-minty problem usually assumes the non-negativity of the parameter vector components. The upper bound $\hat{\bm{y}}$ is set in accordance with the translation in Eq.~\eqref{transf}.}. They define a reasonable limitation of high dimensional search spaces and allow for the generation of initial candidate solution populations.
	    
	Accordingly, the feasible region $M\subset\mathbb{R}^N$ is determined by $2N$ inequality constraints. It forms a perturbed unit hypercube within the subset of the search space that is determined by the box-constraints of each $\bm{y}\in\mathds{R}^N$. The objective function $\bm{c}^\top\bm{y}$ is determined by the vector
	\begin{equation}
	  \label{KMc}
	  \bm{c} = (0, 0, \dots, 0,  1 )^\top \in \mathbb{R}^N.
	\end{equation}
	It is designed in such a way that the optimal solution of the Klee-Minty problems is located at $\bm{y}^* = \bm{0}\in\mathds{R}^N$. 
	The optimal objective function value is $f_{opt} = 0$. 
	
	Considering the design of problem~\eqref{KMopt}, the optimal solution is always located at the origin of the $N$-dimensional search space. 
	This construction can attribute bias in two different ways. On the one hand, the location of $\bm{y}^*$ favors algorithms that predominantly search in direction of the Cartesian axes, e.g. Coordinate Search or certain DE algorithms~\cite{SuttonLW2007}. 
	On the other hand, the non-negativity requirement $\bm{y} \geq 0$ might be handled in a box-constrained approach.~\footnote{In the field of EA, several methods to treat box-constraints do exist, e.g. by random reinitialization inside the box or by repair of violated components.} 
	Hence, the optimal solution $\bm{y}^*$ is located on the boundary of that respective box. In case that a search algorithm is allowed to take into account infeasible candidate solutions in its procreation process, this can have a considerable effect on its performance.
	That is, the creation of infeasible solutions outside the box may be compensated by the box-constraint handling approach. 
	Depending on the method used, the box-constraint handling might bias the search towards infeasible candidate solutions that are repaired in a beneficial way and thus approach the optimal solution more quickly and/or with higher precision.
	
	Both issues are resolved by introducing a direct motion of the vectors in the parameter space.
	The direct motion is an isomorphic transformation that preserves the orientation of a parameter vector $\bm{y}$
	\begin{equation}
	\begin{split}
	  T(\bm{y}):  \mathbb{R}^N \to & \:\mathbb{R}^N \\
		\bm{y} \:   \mapsto & \: \tilde{\bm{y}} = \bm{R} (\bm{y} - \bm{t}).
		\end{split}
		\label{transf}
	\end{equation}
	The transformation consists of two components. A translation by the vector $\bm{t}$ that is followed by a rotation $\bm{R}$ in a suitable hyperplane of the search space.
	The terms $\bm{R} $ and $\bm{t}$ are arbitrarily chosen in the following way.
	The $N$-dimensional rotation matrix $\bm{R}$ is composed of those two orthonormal vectors $\bar{\bm{v}}_1$ and $\bar{\bm{v}}_2$ that span a two-dimensional hyperplane of the search space.
	They read
	\begin{equation}
	\begin{split}
	\bar{\bm{v}}_1 & = ( 0 , \dots,  0 , 1 )^\top  \in \mathbb{R}^N, \quad \textrm{and} \\
	  \bar{\bm{v}}_2 &= \cfrac{\bm{u}}{\left\|\bm{u}\right\|} \: \textrm{ with } \: \bm{u}= (1 , \dots , 1, 0 )^\top \in \mathbb{R}^N
	  \end{split}
	\end{equation}
	The matrix is then build as
	\begin{equation}
		\bm{R} = \bm{I} + ( \cos(\varrho)-1)\left(\bar{\bm{v}}_1 \bar{\bm{v}}_1^\top + \bar{\bm{v}}_2 \bar{\bm{v}}_2^\top\right) - \sin(\varrho) \left(\bar{\bm{v}}_1 \bar{\bm{v}}_2^\top - \bar{\bm{v}}_2 \bar{\bm{v}}_1^\top\right),
	\end{equation}
	where $\bm{I}$ denotes the $N$-dimensional identity matrix and the term $\varrho$ refers to the rotation angle.
	The matrix $\bm{R}\in \mathbb{R}^{N \times N}$ is an orthogonal matrix, i.e. $\bm{R}\bm{R}^\top = \bm{I}$, with determinant $\textrm{det}(\bm{R})=1$. Hence, it represents a rotation about the origin of the $N$-dimensional search space.
	Aiming at a reasonable amount of complexity, the considered rotation angle is preset to $\varrho=\frac{350}{180} \pi$, and the $N$-dependent translation vector\footnote{The choice of the translation $\bm{t}$ represents an empirically motivated compromise between complexity and numerical stability for a wide range of search space dimensions.} $\bm{t} = ( N^3, N^3 , N^3 , \dots ,N^3 )^\top$ is chosen.
	\begin{figure}[!t]
	\centering
	 \includegraphics[width=0.5\textwidth,height=0.37\textwidth]{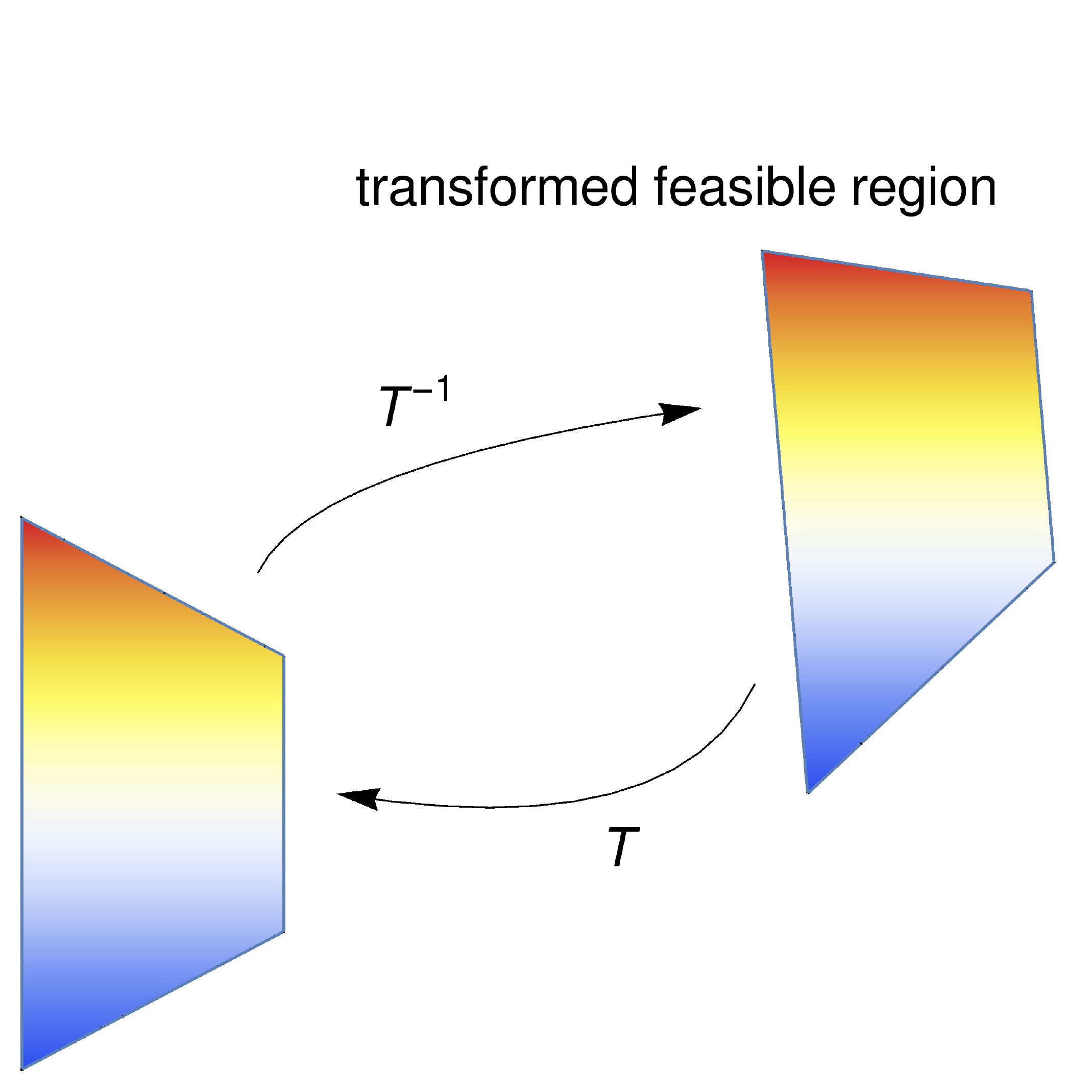}
	 \caption{Translation and rotation of the $2$-dimensional Klee-Minty polytope. The contour lines of the corresponding objective function are indicated by the color (or grey-scale) gradient.}
	 \label{fig01}
	\end{figure}
	
	By application of $T(\bm{y})$, the vertices of the Klee-Minty cube are relocated. Hence, the optimal solution of~\eqref{KMopt} is displaced as well. As the origin is not affected by the rotation, the optimal solution is transferred into $\tilde{\bm{y}}^* \coloneqq T^{-1}(\bm{y}^*)=\bm{t}$. Refer to Fig.~\ref{fig01} for an illustration of the two-dimensional scenario.
	
	While the search is carried out, constraint violation is evaluated after transformation with respect to $T(\bm{y})$. The objective function is left unchanged.
	As a consequence, the Klee-Minty problem~\eqref{KMopt} transforms into 
		\begin{equation}
		\begin{split}
		    \min_{\bm{y}\in \mathbb{R}^N} &\: f(\bm{y}) = \bm{c}^\top\bm{{y}} \\
		      \textrm{s.t.} & \:\bm{A}\bm{R} (\bm{y} - \bm{t}) \leq \bm{{b}},\\
			   &  \:  \check{\bm{y}} \leq \bm{y} \leq \hat{\bm{y}}. 	    
		\end{split}\label{KMoptRot}
		\end{equation}
	with $\bm{c} = (0, 0, \dots, 0,  1 )^\top $.
	Problem~\eqref{KMoptRot} is referred to as \emph{the rotated Klee-Minty problem}. It represents our proposal of a linear constrained benchmark environment that scales the number of linear inequality constraints with the dimension.
	By construction, the optimal objective function value is $F_{opt} = N^3$ for the relocated optimal solution.
	
	While the rotation angle, the rotation plane, as well as the translation vector can essentially be determined randomly,
	this section considers fixed values as a first step.
	Note that the positive orthant is still used to generate an initial population of candidate solutions.
	The relocated optimal solution $\tilde{\bm{y}}^*$ may conceivably still be placed in the positive orthant of $\mathds{R}^N$. 
	To ensure the optimality of $\tilde{\bm{y}}^*$ rotations about angles $\varrho \in [\frac{3}{2} \pi , 2\pi ]$ in the constructed hyperplane are admissible.
	It must be noticed that rotations of the feasible region may simplify the rotated Klee-Minty problem for LP solvers considerably. 
	This is due to the definition of the objective function which is geared to slow progress of the LP solvers when iterating through the Klee-Minty cube.
	However, the focus of this paper is on the comparison of EA variants for constrained optimization. 
	Hence, the rotations appear reasonable to address the rotational invariance of EA strategies.

\section{Benchmarking conventions}
\label{convention}
	In order to use the rotated Klee-Minty problem~\eqref{KMoptRot} as a constrained benchmark for algorithm comparison, 
	some benchmarking principles need to be specified. These aim at providing a comprehensive benchmarking environment that allows for generating reproducible and comparable test results. In any case, algorithm developers intending to use the specified benchmark environment are prompted to carefully report on their complete algorithmic details.	
	
	Considering the original Klee-Minty cube representation~\cite{klee1970good}, problem dimensionalities $N\geq16$ would have to be excluded due to numerical instabilities. Instead, the perturbed unit cube~\cite{deza2006central} presented in Sec.~\ref{kleeminty} allows for the consideration of larger $N$ values.
	The proposed linear constrained benchmark problem takes into account search space dimensions $N\in\{2,3,5,10,20,40\}$\footnote{This a first suggestion; larger search space dimensions can easily be included.}. Accordingly, $6$ distinct constrained functions are considered as benchmark set. 
	
	While the analytical formulation of problem~\eqref{KMoptRot} is available, tested algorithms are expected to treat the problem like a black-box.
	Each evaluation of the whole constrained function is accounted one function evaluation. The predefined budget of function evaluations is $2\cdot10^4N$. However, this represents a first recommendation and may be changed according to choice.
	
	Besides the maximal number of function evaluations, two optional termination criteria are proposed.
	The search may successfully stop after an algorithm approaches the known optimal function value of $N^3$ by a factor of $10^{-8}$.
	Further, algorithms might stagnate in suboptimal edges of the feasible hypercube.
	That is, the search is also terminated after the best-so-far solution is not improved for a predefined number of generations, e.g. $1\%$ of the number of function evaluations ($100 N$). This can save considerable amounts of experimentation time.\footnote{All three termination criteria were considered for both ES variants to realize the experimental results displayed in Sec.~\ref{presentation}. Instead, random search omits the third termination criterion as stagnations are likely.}
		
	Taking into account the definition of problem~\eqref{KMoptRot}, a box-constrained handling technique is dispensable. However, it may be applied to enforce searching in the non-negative orthant of $\mathds{R}^N$.
	For initialization, a starting point (or population) is supposed to be randomly sampled inside the non-negative orthant of $\mathds{R}^N$. 
	Each algorithm should execute at least $15$ independent runs on each constrained function, i.e. in each dimension $ N \in \{2,3,5,10,20,40\}$. 
	
	Final candidate solutions $\bm{y}$, and $\bm{z}$, realized in different algorithm runs are compared by use of a \emph{lexicographic ordering} $\preceq_\textrm{lex}$ that takes into account the objective function value $f(\bm{y})$ as well as the corresponding amount of constraint violation $\nu(\bm{y})$. The respective order relation is defined by
	\begin{equation}
	  \label{lexo}
	      \bm{y} \preceq_\textrm{lex} \bm{z} \Leftrightarrow \left\{ \begin{matrix}
				      f(\bm{y}) \leq f(\bm{z}), & \quad \textrm{if }\nu(\bm{y}) =\nu(\bm{z}),\qquad \\
				      \nu(\bm{y}) < \nu(\bm{z}), & \textrm{else}.
				\end{matrix}  \right.
	\end{equation}
 	Hence, in the context of the rotated Klee-Minty problem, the objective function value corresponding to $\bm{y}$ is $f(\bm{y}) \coloneqq \bm{c}^\top \bm{y}$. The constraint violation value $\nu(\bm{y})$ can be measured as the sum of the deviation over all inequality constraints\footnote{It is recommended to use the mentioned constraint violation definition. As there exist multiple different ways to define the constraint violation, algorithm developers may use their definition of choice. However, a detailed explanation is obligatory to ensure the comparability of algorithm test results.}
	\begin{equation}
	  \label{convio}
	  \nu(\bm{y}) \coloneqq \sum_{i=1}^N{\max\big\{0,\big(\bm{A} \bm{R} \bm{y}-\bm{A} \bm{R}\bm{t}-\bm{b}\big)_i\big\}}.
	\end{equation}
	The lexicographic order relation permits to define a number of quality indicators that can be used to assess and compare algorithm performance on  problem~\eqref{KMoptRot}. 
	\begin{table}[t]
	\centering\caption{Quality indicators for algorithm assessment and comparison on the rotated Klee-Minty problem. The indicators refer to the results obtained from $15$ independent algorithm runs and make use of the ordering relation  according to $\preceq_\textrm{lex}$ in Eq.~\eqref{lexo}.  }
	\renewcommand{\arraystretch}{1.1}
	\begin{tabular}{p{2.4cm}p{8.5cm}}\hline
	 $f_\textrm{best}$ & fitness of the best found solution \\
	 $f_\textrm{med},\nu_\textrm{med}$ & fitness and constraint violation of the median solution\\
	 $|f_\textrm{med} - f_\textrm{opt}|$ & absolute error of the median solution \\
	 $FR$ & feasibility rate $FR = \frac{\# \textrm{feasible runs}}{\# \textrm{runs}}$\\
	 $\left\|\bm{y}-\tilde{\bm{y}}^* \right\|$   & mean deviation of all final and feasible algorithm realizations $\bm{y}$ from the known optimal solution $\tilde{\bm{y}}^*$ over $15$ algorithm runs \\
	 meanFevals & mean number of function evaluations until termination\\
	\end{tabular}
	\label{KMqual}
	\end{table}
	
\section{Algorithm assessment and comparison}
	\label{presentation}
	This section is concerned with the evaluation and presentation of the algorithm results obtained in $15$ independent runs on the rotated Klee-Minty problem~\eqref{KMoptRot}. 
	To this end, two EA variants are exemplarily tested and compared to Random Search (RS). In particular, we consider the Differential Evolution (DE) variant LSHADE44~\cite{PolakovaT2017} (CEC2017 competition winner) and the Evolution Strategy (ES) for constrained optimization which is called $\epsilon$MAg-ES\footnote{For performance improvements, the $\epsilon$ threshold is initially set to zero in all algorithm runs, i.e. the $\epsilon$-level ordering is replaced with the lexicographic ordering~\eqref{lexo}.}~\cite{HB2018a}.
	
\begin{figure}[htbp]
	\centering
	 \includegraphics[trim= 30 180 60 230,clip,width=0.45\textwidth,height=0.335\textwidth]{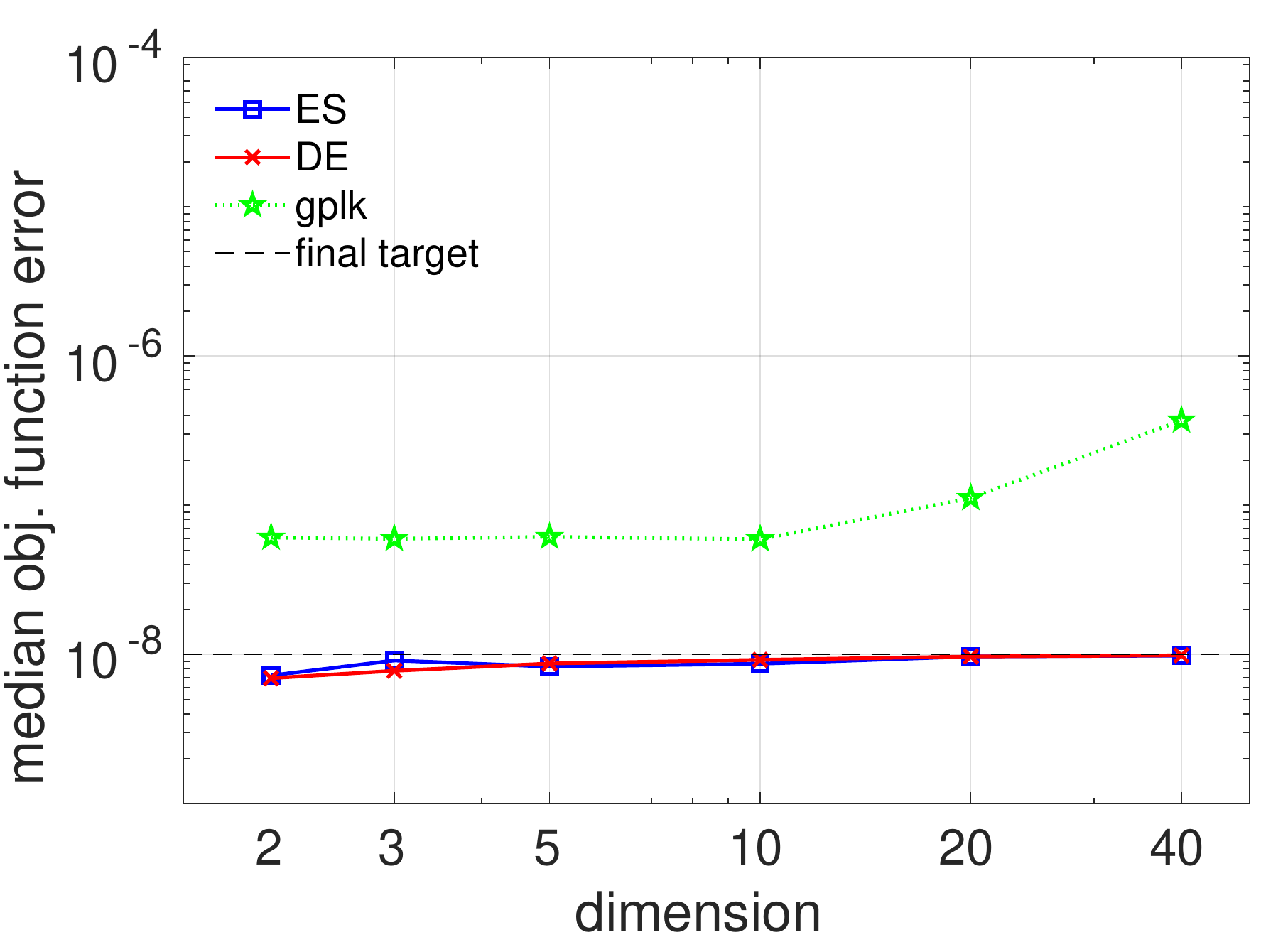}\qquad
	 \includegraphics[trim= 20 180 60 230,clip,width=0.45\textwidth,height=0.33\textwidth]{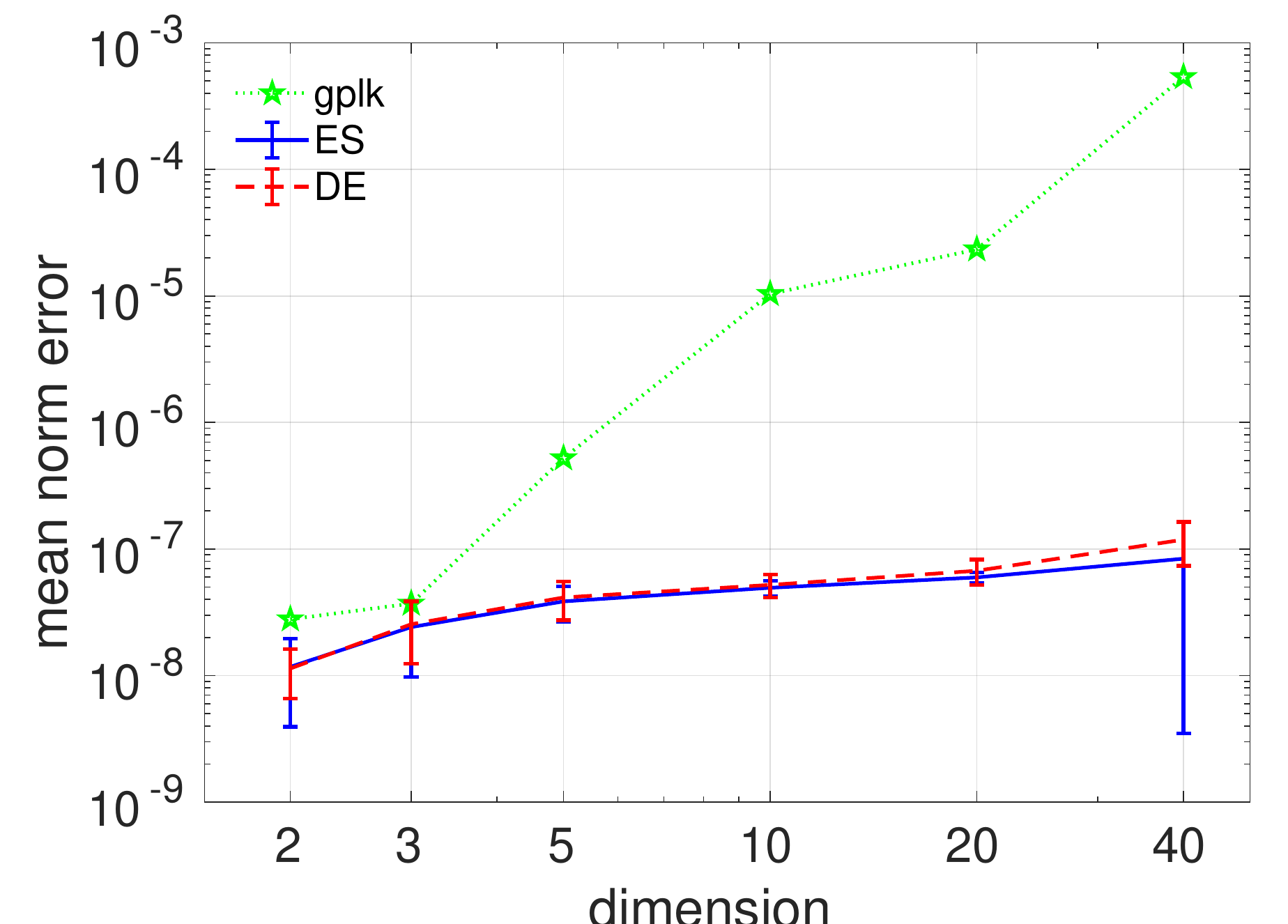}
	 \caption{Illustration of the algorithm performance with respect to the quality indicators $|f - f_\textrm{opt}|$ and $\left\|\bm{y}-\tilde{\bm{y}}^* \right\|$ reported in Table~\ref{KMtab1}. All reports are plotted against the search space dimension.}
	 \label{fig02}
	\end{figure}

	In order to make a statement about algorithm performance, different quality indicators are derived from the $15$ final algorithm realizations.
	Table~\ref{KMqual} specifies these indicators. Accordingly, taking into account different search space dimensions, the final results are presented in the form of Table~\ref{KMtab1}.
	 This presentation style is inspired by the CEC competitions on constrained real-parameter optimization~\cite{CEC2017}. It allows comparing different algorithms with respect to realized median objective function values. Further, algorithm performance in the search space is measured by taking into account the mean deviation of the best found candidate solution from the optimal solution.	
	\begin{table}[t]
	\centering\caption{Results of $\epsilon$MAg-ES and LSHADE44 on problem~\eqref{KMoptRot} in dimension $2$ to $40$.}
	 \scalebox{0.88}{
	 \begin{tabular}{p{0.8cm}p{0.8cm}p{1.8cm}p{1.8cm}p{1.2cm}p{1.8cm}p{0.8cm}p{1.8cm}p{1.8cm}}
	  \multicolumn{9}{c}{$\epsilon$MAg-ES} \\\hline \hline
	  $N$ & $f_\textrm{opt}$ & $f_\textrm{best}$ & $f_\textrm{med}$ &  $\nu_\textrm{med}$ & $|f_\textrm{med} - f_\textrm{opt}|$  & $FR$ & $\left\|\bm{y}-\tilde{\bm{y}}^* \right\| $ & meanFevals \\ \hline
	  $2$ & $2^3$ & 8.0000e+00 & 8.0000e+00 & 0 & 9.7490e-09  & 1.00 &2.0549e-08&  1.5506e+03 \\ 
	  $3$ &  $3^3$ & 2.7000e+01 &2.7000e+01 & 0 & 7.5230e-09  & 1.00 &1.0970e-08 & 4.2336e+03 \\ 
	  $5$ &  $5^3$ & 1.2500e+02& 1.2500e+02 & 0 & 8.7761e-09  & 1.00 & 3.5589e-08& 1.631e+04 \\ 
	  $10$ &  $10^3$& 1.0000e+03& 1.0000e+03 & 0 & 8.8155e-09  & 1.00 & 4.6960e-08& 2.6747e+04 \\ 
	  $20$ &  $20^3$ &8.0000e+03 & 8.0000e+03 & 0 & 9.8480e-09  & 1.00 &5.7747e-08 & 8.5109e+05 \\ 
	  $40$ &  $40^3$ & 6.4000e+04& 6.4000e+04 & 0 & 9.8225e-09  & 1.00 &   8.3878e-08    & 3.4478e+05 \\ 
	  \\
	    \multicolumn{9}{c}{LSHADE44} \\\hline \hline
	  $N$ & $f_\textrm{opt}$ & $f_\textrm{best}$ & $f_\textrm{med}$ &  $\nu_\textrm{med}$ & $|f_\textrm{med} - f_\textrm{opt}|$  & $FR$ & $\left\|\bm{y}-\tilde{\bm{y}}^* \right\| $ & meanFevals \\ \hline
	  $2$ & $2^3$ & 8.0000e+00 & 8.0000e+00 & 0 & 7.6762e-09  & 1.00 &1.8423e-08&  3.9534e+03 \\ 
	  $3$ &  $3^3$ & 2.7000e+01 &2.7000e+01 & 0 & 8.9507e-09  & 1.00 & 2.4902e-08& 8.4597e+03 \\ 
	  $5$ &  $5^3$ & 1.2500e+02& 1.2500e+02 & 0 & 9.4049e-09  & 1.00 & 4.5918e-08 & 2.3171e+04 \\ 
	  $10$ &  $10^3$& 1.0000e+03& 1.0000e+03 & 0 & 9.4270e-09  & 1.00 & 5.0702e-08 & 7.9120e+04 \\ 
	  $20$ &  $20^3$ &8.0000e+03 & 8.0000e+03 & 0 & 9.7224e-09  & 1.00 & 1.1357e-07& 2.1813e+05 \\ 
	  $40$ &  $40^3$ & 6.4000e+04& 6.4000e+04 & 0 & 2.8513e-09  & 1.00 & 1.1861e-07 & 5.7090e+05 \\ 
	  \end{tabular}
	  }	    	  
	   \label{KMtab1}
	\end{table}	
	As lengthy and hardly comparable tables should ideally be supported with easily interpretable figures~\cite{Johnson2002}, we provide an illustration of these information in Figure~\ref{fig02}.  It can be observed that both EA variants approach the optimal objective function value with the requested precision up to dimension $N=40$. Compared to the results of the Interior Point LP solver \texttt{glpk}~\ref{KMtab2}, both EA variants obtain solutions of improved quality in terms of objective function values and parameter vector accuracy.
	   \begin{table}[b]	
	\centering \caption{Results obtained by the deterministic Octave interior point LP-solver \texttt{gplk}.}

	 \scalebox{0.88}{
	\begin{tabular}{p{0.8cm}p{0.8cm}p{1.8cm}p{0.8cm}p{1.8cm}p{0.8cm}p{1.8cm}p{1.8cm}}

	\hline\hline
	  $N$ & $f_{\textrm{opt}}$ & $f$ & $\nu$ & $|f - f_\textrm{opt}|$  & $FR$ & $\left\|\bm{y}-\tilde{\bm{y}}^* \right\| $ & meanFevals \\ \hline
	$2$ & $2^3$ & 8.0000e+00 & 0 &  9.7600e-09 & 1.00& 2.7862e-08 & --  \\
	  $ 3$ & $3^3$ & 2.7000e+01 & 0 &  9.2953e-09 &1.00& 3.7072e-08& -- \\
	  $ 5$ & $5^3$ & 1.2500e+02 & 0 &  1.0744e-07 &1.00& 5.2181e-07& -- \\
	  $ 10$ & $10^3$ & 1.0000e+03 & 0 &  1.9369e-06 &1.00& 1.0368e-05& -- \\
	  $ 20$ & $20^3$ & 8.0000e+03 & 0 &  4.1322e-06 &1.00& 2.3340e-05& -- \\
	  $ 40$ & $40^3$ & 6.4000e+04 & 0 &  9.5593e-05 &1.00& 5.3734e-04 & -- 

	\end{tabular}	
	  }	 
	  \label{KMtab2}
	\end{table}
	
\begin{figure}[t]
	  \centering
	  \includegraphics[trim= 55 220 65 230,clip,width=0.95\textwidth]{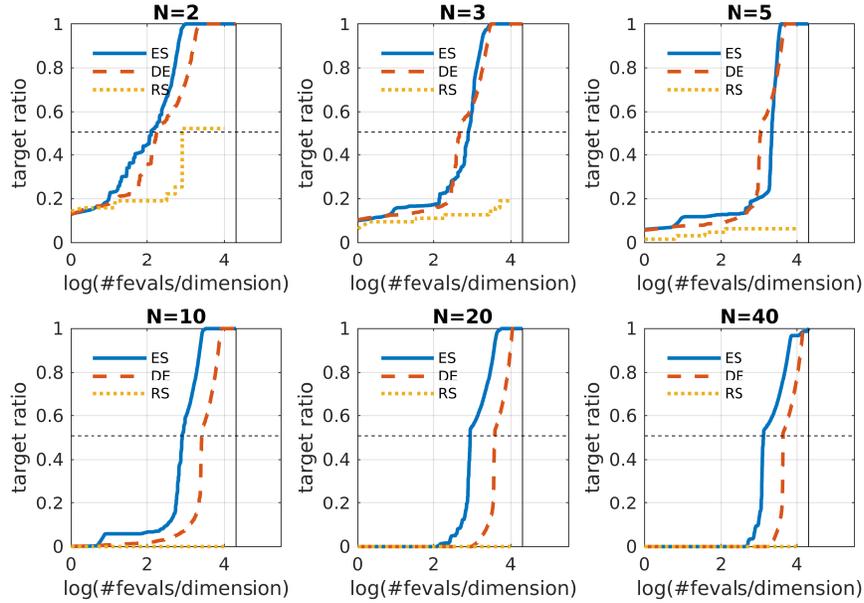}
	  \caption{ECDF results of the $\epsilon$MAg-ES (solid lines) and the LSHADE44 (dashed lines) on problem~\eqref{KMoptRot}. The dotted lines show the baseline results of Random Search (RS), respectively.  }
	  \label{fig03}
	\end{figure}

		While the CEC2017 competition does not include a notion of runtime into the algorithm assessment, we address this issue in two ways. On the one hand, the actual number of constrained function evaluations need to be reported, see Table~\ref{KMtab1}. On the other hand, performance profiles or empirical cumulative distribution function (ECDF) plots are introduced.  The runtime of meta-heuristic algorithms can be directly identified with the number of function evaluations needed to satisfy a number of predefined targets. This runtime definition can be traced back to~\cite{MoreWild2009} and is widely used in the context of the COCO BBOB benchmarks\footnote{A more detailed explanation of the ECDF construction and interpretation is provided in~\cite{Hansen2016perf}. Notice, this version of the rotated Klee-Minty benchmark omits the use of the bootstrapping approach mentioned in that respective paper.}. However, the target definition used in this paper is different.
			Instead of defining targets only for the feasible region of the search space, we allocate $50\%$ of the targets to the infeasible region.
	This supports the illustration of algorithm behavior within the infeasible region. Refer to Figure~\ref{fig03} for a demonstration.
	In total $103$ target values are defined. The $51$ targets in the infeasible region are uniformly distributed between $10^{4}$ and $10^{-6}$ as well as $0$.
	
	Realizing a candidate solution with constrained violation below a target definition for the first time, a target value is considered to be hit.
	The targets in the feasible region range from $10^0$ to $10^{-8}$. They are reached after having realized a feasible candidate solution with an objective function value smaller than a certain target value. The ECDF plots display the ratio of reached targets for any number of function evaluations. This way they provide a notion of algorithm performance: Smaller upper left areas indicate faster algorithm running times~\cite{MoreWild2009}.
	Both runtime illustrations (in Table~\ref{KMtab1} and in Figure~\ref{fig03}) display the runtime advantage of the $\epsilon$MAg-ES for search space dimensionalities $N\leq40$. Its advantage on the rotated Klee-Minty problem appears to grow with the search space dimension. Note that this runtime definition aims at the comparison of probabilistic search algorithms. Hence, it is not compatible with a comparison to the LP solver.
	
	The algorithms are compared in each individual dimension. We propose to rank two algorithms according to the quality indicators displayed in Figure~\ref{fig02} as well as their run times illustrated in Figure~\ref{fig03}. The use of three distinct equally weighted ranking factors avoids ties. In this respect, the ES receives a better rank than LSHADE44 as it basically realizes solutions of similar quality but revealing faster running times in terms of the function evaluations needed to reach the given targets.
	However, the final design of the ranking procedure is not ultimately determined.  
	After having ranked all algorithms in every dimension, an overall winner can be determined by aggregating over all dimensions if considered necessary.
	
	The developmental stage of a Matlab implementation of the introduced rotated Klee-Minty benchmarking environment is made publicly available in a Github repository\footnote{\href{https://github.com/hellwigm/RotatedKleeMintyProblem}{https://github.com/hellwigm/RotatedKleeMintyProblem}}.
	
	\section{Conclusion}
	\label{concl}
	
	This paper suggests a novel set of linear constrained optimization problems that are suitable for benchmarking probabilistic search algorithms in a black-box setting. To this end, the Klee-Minty problem known from linear programming was modified. The emerging optimization problem is referred to as rotated Klee-minty problem. Further, corresponding reporting and presentation rules are specified to ensure reproducible and comparable benchmarking results.
		{Still, the benchmark problems are rather statically designed.}
	A first recommendation for the ranking of competing algorithms is provided. However, additional investigations are necessary to decide whether the proposed consensus ranking can be improved. 
	Furthermore, the introduction of redundant constraints reduces the performance of interior point methods (without preprocessing steps) considerably~\cite{deza2006central}. As the EA performance is considered independent of such constraints, their incorporation into the proposed black-box benchmark definition may be a task for future research.
	These tasks and other advancements of the benchmark environment will be addressed in future research.	
	
	Representing a first proposal of a Klee-Minty based black-box optimization benchmark environment for EA variants, a constructive discussion of the benchmark design is welcome. Please contact us with suggestions for improvements and/or modifications.
     
\section*{Acknowledgements}
The work was supported by the Austrian Science Fund under grant P29651-N32.

\bibliographystyle{splncs04}

\end{document}